# Can generative AI figure out figurative language? The influence of idioms on essay scoring by ChatGPT, Gemini, and Deepseek

Enis Oğuz

Middle East Technical University, Turkey

## Abstract

The developments in Generative AI technologies have paved the way for numerous innovations in different fields. Recently, Generative AI has been proposed as a competitor to AES systems in evaluating student essays automatically. Considering the potential limitations of AI in processing idioms, this study assessed the scoring performances of Generative AI models for essays with and without idioms by incorporating insights from Corpus Linguistics and Computational Linguistics. Two equal essay lists were created from 348 student essays taken from a corpus: one with multiple idioms present in each essay and another with no idioms in essays. Three Generative AI models (ChatGPT, Gemini, and Deepseek) were asked to score all essays in both lists three times, using the same rubric used by human raters in assigning essay scores. The results revealed excellent consistency for all models, but Gemini outperformed its competitors in interrater reliability with human raters. For essays with multiple idioms, Gemini followed a the most similar pattern to human raters. While the models in the study demonstrated potential for a hybrid approach, Gemini was the best candidate for the task due to its ability to handle figurative language and showed promise for handling essay-scoring tasks alone in the future.

**Introduction**

Artificial Intelligence (AI) has revolutionized numerous fields in recent years, increasing the need for interdisciplinary collaborations with AI technologies (Sengar et al. 2024). Generative AI models have been applied to various tasks across a wide range of disciplines, including engineering, linguistics, and education. Although the integration of AI technologies into linguistics and education has raised some concerns, overall results suggest significant improvements in saving time and resources. Recently, the question has arisen whether generative AI can be used for scoring essays, which is a demanding and time-consuming task in education. One major concern for such an application is the rather limited ability of generative AI models to interpret figurative language, particularly idioms. If generative AI models struggle to comprehend idiomatic expressions, the score assigned for an essay might tend to become less reliable by the increasing number of idioms. This would create a notable disadvantage against students who use figurative language and idioms frequently, despite their success in using one of the most challenging aspects of language. Addressing this potential shortcoming, this study investigated the influence of idiomatic expressions on the essay-scoring performance of three generative AI models (ChatGPT, Gemini, and DeepSeek), along with the consistency of each model in evaluating the same essays.

**Literature Review**

The involvement of AI in linguistics and education has both positive and negative potential outcomes (Kasneci et al. 2023). On the negative side, generative AI can become a plagiarism tool by completing writing assignments and projects on behalf of students (Perkins, 2023). After all, recent AI models can produce academic assignments with a level of depth and accuracy that is even acceptable in higher education (Williams, 2024). This raises concerns regarding academic integrity and student development, and institutions try to avoid such misuses by seeking plagiarism-detecting tools and trying to educate students on ethical principles. On the positive side, AI can act as an assistant and provide feedback and suggestions to improve academic skills (e.g., Nazari et al., 2021). AI feedback is unique in its ability to provide a tailored approach for each and every student, creating a personalized and efficient experience. However, the potential benefits of AI have not yet been fully explored, especially across different assessment types and academic disciplines (Williams, 2024).

Writing assessment is one particular area that can greatly benefit from AI. Reliable scoring of essays and other writing tasks is difficult due to the significant costs of money and time to train required raters (Meyer et al., 2023). Holistic scoring of such tasks can bring time-efficiency but also high subjectivity and less reliability, compared to more objective and reliable rubric-based assessment (Yavuz et al., 2025). The dilemma, then, lies in choosing between efficiency and reliability, as improving one harms the other. As an alternative approach, AI can offer the best of both worlds; in terms of time efficiency, AI-based assessment can evaluate hundreds or even thousands of essays within seconds, and unlike traditional essay scoring, this efficiency does not cost reliability.

The integration of AI in essay scoring goes back to the 1960s, with the introduction of Automated Essay Scoring (AES) systems (Page, 1966). Since then, these systems have been developed at a rapid pace thanks to the interdisciplinary contributions of computer science, linguistics, and instructional technology (Gierl et al., 2014). Although AES tools have received some criticism (e.g., Condon, 2013), their potential is well-acknowledged in general (Stevenson & Phakiti, 2014). Such tools can reduce the workload and provide reliable assessment (Kumar & Boulanger, 2021), increase the efficiency of teacher feedback (Escalante et al., 2023), and become topic-aware with the integration of language models like BERT (Wu et al., 2023). With some technical knowledge, educators and researchers can use Natural Language Processing (NLP) tools to get an automatic evaluation of features regarding lexical sophistication (Kyle & Crossley, 2014), text cohesion (Crossley et al., 2016), grammatical and mechanical errors (Crossley et al., 2019) and more. These values can be fed into machine learning algorithms, allowing consistent AES systems to be built for any writing task. Recently, a further development in improving such systems has been the integration of deep learning, which can get required features from the data directly without the assistance of humans (Mizumoto & Eguchi 2023). Although this approach does not need any human participation in extracting and analyzing data, the creation of the system requires advanced technical knowledge in machine learning, deep learning, and statistics. For this reason, a deep learning integrated AES system requires significant expertise, workload, and resources.

The rise of Generative AI has immediately raised the question of whether it can perform as well as -or better than- traditional AES systems. The strong desire for an answer stems from the minimal technical knowledge required to use most Generative AI models, unlike the expertise needed to develop a deep learning system. The transformative nature of Generative AI models allows utilizing neural networks in producing text and other forms of media in response to prompts, earning the title "generative" (Williams, 2024). These AI systems are based on large language models (LLM), utilizing machine learning principles and massive datasets. Compared to BERT, one of the largest masked language models with 350 million parameters, the Generative AI model GPT-3 possesses 175 billion parameters (Mizumoto & Eguchi 2023). This increase is exciting given that BERT has already been shown to improve AES tasks even with limited data (Xue et al., 2021). Recent experimental studies have revealed promising results concerning the use of Generative AI in language-related tasks. For instance, GPT has been shown to produce better argumentative essays than humans (Herbold, et al., 2023) and outperform crowd workers in text annotation tasks (Alizadeh et al., 2023), challenging the necessity of AES systems in achieving human-like or even better performance in language-related tasks. In particular, the effectiveness of Generative AI in automated essay scoring has recently been a hot debate, with comparisons made to human raters and AES performance.

The findings are mixed when it comes to Generative AI models and essay scoring. Some studies argue that Generative AI models, such as ChatGPT and Gemini, can provide acceptable measures of student essay quality (e.g., Mizumoto & Eguchi, 2023), while others emphasize their limitations on such tasks (e.g., Bui & Barrot, 2024; Manning et al., 2025). One area where Generative AI thrives is consistency. GPT 4, for instance, can reach about 80% consistency in scoring, compared to 43% for human raters; however, this consistency might also be due to AI's avoidance of giving scores on the edges, namely, the lowest or the highest scores (Tate et al., 2024). The high consistency of AI was confirmed in another study, in which both ChatGPT and Bard (the precedent of Gemini) achieved remarkable consistency in scoring three different essays (Yavuz et al., 2025). While the limited number of essays in this last study necessitates further investigations, the lack of conclusive evidence regarding consistency is not the only limitation Generative AI has. Overall, Generative AI seems to fall behind state-of-the-art solutions, performing at an average level across tasks without excelling in any of them (Kocoń et al., 2023). That is why although Generative AI models are praised for their extensive feedback and acceptable scoring performance, a hybrid method has been proposed as a better alternative, as AI might fail to show a strong correlation with human raters (e.g., Khademi, 2023; Mizumoto & Eguchi 2023). In the end, the integration of AI would be pointless if it cannot perform as well as human raters, and pairing such models with experienced raters would function as a safety switch, allowing a third rater to be involved in cases of significant scoring differences. Nevertheless, the ability of Generative AI as a second rater still needs further confirmation.

One challenging area for Generative AI is comprehending figurative language. Even human experts may struggle with translating figurative expressions across languages, as this task requires more than simply replacing translation equivalents (Sahari et al., 2024). ChatGPT has demonstrated adequate performance in translation tasks, but the efficiency dropped with low-resource or distant languages (Hendy et al., 2023; Jiao et al., 2023), raising the question of whether such models can achieve the same performance when the data is limited. In particular, figurative language use is less frequent than literal language use, especially in academic contexts, and the limited availability of figurative expressions can lead to a lower performance of Generative AI systems in processing them. This argument has received empirical support in the literature as well. Despite the impressive performance of Generative AI models in scoring metaphor creativity (DiStefano et al., 2024), they can struggle with analyzing figurative language in translation and fall behind human translators, especially when idioms are involved (Olivero, 2024; Sahari et al., 2024). In addition to the scarcity of idiomatic expressions in academic texts, the advanced reasoning they require can also be behind this limitation, in line with the challenge second language learners experience with idioms in another language (Beck & Weber, 2016). Unfortunately, the question of whether such a limitation would also manifest itself in automatized essay scoring has remained largely unanswered.

This study investigated the reliability of Generative AI models in essay scoring when idioms are involved. Using a large-scale corpus of student texts, AI models were used to score hundreds of essays, with the number of idiomatic expressions ranging from fourteen to none. In addition to the total number of idioms, a unique idiom count for each essay was also considered by creating an idiom repetition variable. The AI models (GPT 4, Gemini, and Deepseek) were asked to score essays using the same rubric human raters had used, and each essay was scored by each model three times on a different day, allowing an investigation of both internal consistency and inter-rater consistency with human raters. The results revealed important insights regarding the application of Generative AI to essay scoring and other language-related tasks involving figurative language.

**Method**

The current study adopted an empirical design. The essays used in the analyses were extracted from the PERSUADE 2.0 corpus (Crossley et al., 2024). The corpus had over 25000 argumentative essays written by 6th – 12th grade students, who were native speakers of English in the United States of America. Using a holistic rubric, trained raters scored the essays using a double-blind rating process, and expert raters also served as third raters to ensure scoring reliability. This process allowed reliable ratings to be attached to each essay in the corpus.

To determine the number of idioms in essays, an idiom list was created using two sources: the Oxford Dictionary of English Idioms (Ayto, 2010) and Miller (2019). The total number of idioms in the list was 1583. Using R (R Core Team, 2023), a script was created to look for these idioms throughout the essays. First, the script standardized all pronouns, possessives, and articles in the essays and idioms. Punctuation, white spaces, and numbers were also removed, and all letters were transformed into lowercase. Finally, all suffixes were removed from both the idiom list and essays. The procedures ensured a better matching performance in finding idioms. The phrases marked as idioms were stored in a list along with their matched idioms and essay ID numbers. 12417 potential idioms were found. After examining the list, false matches were deleted, and some uncertain entries were checked in the essays. The remaining dataset had 10384 matched idioms. In the next step, a total idiom count and a unique idiom count were calculated for each essay.

174 essays with the most unique idioms (range: 4-7) and the most total idioms (range: 4-14) were selected for the idiomatic essay list. To create a matching control, another R script was created, which found essays without any idioms that had the same given scores and nearly identical word counts. The process ensured that the essays were of similar quality and word count between the idiomatic essay and control lists, allowing an investigation of idiomatic expression's potential influence on given AI scores. The idiomatic essay list had an average score of 4,41 and an average word count of 593,94, while the control essay list had the same average score and almost identical word count average (595,50).

In the final step of data collection, each Generative AI was given the rubric ons: used in scoring the original data and asked to score the essays following the rubric. API services of ChatGPT (gpt-4o) and Gemini (1.5 Pro) were used, and the web-based version of DeepSeek (V3) was used as API service was unavailable at the time. This procedure was repeated for three consecutive days to evaluate the consistency of the models as well. The given scores were added to the final datasets, along with the name of the Generative AI used to associate the particular score. The final dataset included 348 essays, associated scores by human raters and Generative AI models.

The study aimed to answer following research questions:

1) How consistent are Generative AI models in scoring student essays?

2) To what extent do Generative AI models provide scores that align with those given by human raters in essay scoring tasks?

3) Does the presence of idioms in student essays decrease the reliability of scores given by Generative AI?

4) How close are the scoring patterns of Generative AI and human raters across varying number of idioms in essays?

**Data Analysis**

The data was stored in a long format with the following columns: essay ID, essay score (given by human raters), word count, unique idioms (the number of unique idioms in each essay), total idioms (the number of total idioms in each essay), ai score (scores given by AI tools), ai (the tool used for scoring), and analysis number (as the same essays were scored three times by each tool). Analyses were carried out in R (R Core Team, 2023), plots were visualized with the ggplot2 package (Wickham, 2011), models were created with the mgcv (Wood & Wood, 2015) and lme4 (Bates et al., 2015) packages, and ICC values were calculated by the psych package (Revelle & Revelle, 2015).

**Results**

Overall scores showed that Generative AI models generally gave lower scores to essays compared to human raters (Figure 1). While Gemini and ChatGPT showed a similar variety in score distribution compared to human raters, DeepSeek seemed to give scores accumulated around 4 points, showing a rather limited variety. This limited-variety pattern can allow high reliability across different measurements and improve consistency, but it is also likely to harm the inter-rater reliability between the AI model and human raters.

[Insert Figure 1]

The consistency of AI models was investigated using the intraclass correlation coefficient (ICC 3,1) for all three scoring rounds. The results revealed excellent agreement among all three AI models, with DeepSeek reaching the highest consistency with an ICC value of 0.807. Considering the similar values for GPT (0.753) and Gemini (0.796), all models can be regarded as excellent tools in terms of scoring consistency. In order to investigate the consistency between different measurements, the scoring rounds were paired as 1-2, 1-3, and 2-3. The results again showed high consistency across different measurement pairs (Figure 2a), indicating that the models achieved remarkable consistency regardless of the time interval between measurements (1-3 days in the current study). However, as mentioned above, the reason for the best performance of DeepSeek might actually stem from its rather limited scoring variety.

**[Insert Figure 2]**

In the next step, the inter-rater reliabilities between human raters and AI models were calculated, again with ICC (3,1) measurements. Gemini thrived in this area, reaching a notable ICC value of 0.735. DeepSeek reached a value of 0,695, followed by the 0,673 ICC value of ChatGPT. Investigating the issue across scoring rounds revealed a similar picture, with Gemini performing the best and ChatGPT performing the worst in giving similar scores to human raters. Figure 2b shows the inter-rater reliability between AI models and human raters across scoring rounds.

After examining general scoring tendencies and intra-rater and inter-rater reliabilities of Generative AI models, the potential effect of idioms in essays was investigated using a Generalized Additive Model [1] (GAM). The initial idea was to add the total number of idioms as an independent variable to the model. This, however, ignored the potential influence of essay length; an essay with 200 hundred words and 8 idioms would be identical in terms of idiomatic phrases to an essay with 100 words and the same number of idioms. As an alternative solution, the value of the total number of idioms was divided by the word count of each essay and multiplied by 100 (for better visuals), creating a normalized idiom count. The GAM model was created using the following code:

```
model_gam <- gam(ai_score_difference ~ word_count + ai_model +
s(total_idiom_count_normalized, by = ai, bs = "cs") + s(essay_id, bs =
"re"), data = dataset_longformat, method = "REML")
```

---

[1] An investigation of idiom score differences and total idiom number per essay suggested a non-linear relationship between two variables, a decreasing trend at the beginning followed by a steady rise as the number of idioms increases. Therefore, a GAM model was used instead of a Linear Mixed Effects Model.

The model accounted for essay_id as a random effect, with the score difference between Generative AI and human scores taken as the dependent variables. Two independent variables (word count and AI model) and one interaction (total number of idioms normalized x AI model) were included in the model. The GAM model was compared to a linear model; the AIC value was lower for the GAM model, indicating a better fit, and the effective degree of freedom (EDF) of this model was 7.6, suggesting strong nonlinearity. Therefore, the GAM model was used instead of a linear mixed model. Table 1 shows the results of the model.

[Insert Table 1]

Although word count was significant in the model, the effect size was small. All Generative AI models differed significantly from human scores, as the scores assigned were overall lower. The interactions between Generative AI models and the total number of idioms (normalized) suggested an effect of idiomatic phrases on the score differences between AI models and human raters. This potential effect is investigated further below.

Using ICC (3,1), the inter-rater reliability between AI models and human raters was calculated across three measurement rounds for the essays without any idioms in the dataset (Figure 3b). Gemini showed an excellent performance in assigning similar scores to human raters when no idioms were present in the essays, outperforming both DeepSeek and ChatGPT. Next, the same values were calculated for the essays with idioms. Gemini again outperformed the other AI models, this time with a much higher margin (Figure 3a). The results indicated that Gemini was the most similar model to human raters whether idioms were present or not.

[Insert Figure 3]

To better capture the systematic differences between human scores and AI model scores, Figure 4 was created based on a GAM model. The figure showed an initial rise in human scores after idioms were present in essays, followed by a decrease as the number of idioms increased further. Including idioms in an essay increased the given scores, but this positive influence lost its effect and resulted in a detrimental effect when more idioms were used. The overall picture then pointed to a potential negative influence of idiom repetition on essay scores. Although AI models revealed similar patterns, there were important differences as well. Other than the systematically lower scores given by AI models, Gemini followed a very similar pattern to human raters until the total number of idioms peaked. After a certain threshold, Gemini gave higher scores to the essays with more idioms, even surpassing human raters who overall provided higher scores. Deepseek and ChatGPT seemed to follow a less nuanced pattern unlike human raters and Gemini, with Deepseek continuing to decrease scores in line with human raters even in essays with many idioms.

[Insert Figure 4]

The total number of idiom values was normalized in the analyses and figures above by dividing the value by word count. Although this index was sensitive to essay length, it did not consider the potential influence of idiom repetition; repeating the same idioms in an essay would inflate the total idiomatic phrases, which will be different from using unique idioms. In order to alleviate this potential confounding effect, an idiom repetition value was created by using the following formula:

$$\frac{The\ number\ of\ total\ idioms\ /\ The\ number\ of\ unique\ idioms}{Word\ count} \times 1000$$

This index helped account for the overuse of the same idiom. Assuming a word count of 100, using the same four idioms in a single essay once would produce an idiom repetition score of 1, while using each idiom twice would result in an idiom repetition score of 2. Therefore, the increase in this index showed an increase in the repetition of the same idioms in an essay. Figure 5 was created using a GAM model to show how given essay scores were distributed across idiom repetition values. Similar to Figure 4 above, an initial increase was followed by a steady decrease. Although its scores were lower in general, Gemini mimicked the human scoring pattern best, except for the highest idiom repetition values. ChatGPT and DeepSeek showed more straightforward scoring patterns, lacking the variety of human raters and Gemini. Overall, using idioms in a nonrepetitive way increased the scores given by both Generative AI models and human raters, but the scores decreased as idiom repetition increased.

**[Insert Figure 5]**

Finally, inter-rater reliability between human raters and AI models was examined across the number of unique idioms in essays. The interrater reliability stayed stable across the number of unique idioms [2] (except for the decrease in inter-rater reliability for Deep and GPT at the end), confirming the negative influence on idiom repetition (Figure 6). In summary, human scores showed high interrater reliability with Gemini and acceptable interrater reliability with ChatGPT and DeepSeek when the number of unique idioms was considered.

---
[2] Since there was only one essay with 7 unique idioms, it was excluded in the figure.

[Insert Figure 6]

Discussion

This study examined the influence of idioms on essay-scoring reliability and consistency of Generative AI models. AI models gave systematically lower scores to essays compared to human raters. While all three models showed good consistency across three scoring rounds, Gemini provided the best interrater reliability with human raters, especially for essays with idioms. Scoring patterns of human raters showed an initial rise in scores with the increasing number of idioms, followed by a steady decrease due to the potential effect of idiom repetition. Although AI models seemed to follow a similar pattern, ChatGPT and DeepSeek exhibited a less varied scoring pattern, failing to mimic the prominence and variability of the scoring curves provided by Gemini and human raters. Gemini also showed the best inter-rater reliability with human raters across a range of unique idioms in essays.

    The consistency of Generative AI models reached excellent levels across all three measurements, confirming previous studies (e.g., Tate et al., 2024; Yavuz et al., 2025). Although DeepSeek gave the best results for consistency (i.e., intra-rater reliability), it also avoided giving scores on extreme ends, similar to the finding of Tate et al. (2024) for ChatGPT. Given that it had been more than a year since the analyses of that earlier study (March 23, 2023), ChatGPT might have improved its algorithms, allowing a more varied scoring procedure that captured the finer details in the essays examined in the current study. As the majority of scores given by DeepSeek accumulated around 4, the consistency of this model might be misleading, but since

this model was developed years later than ChatGPT, it can also improve its algorithms over time and provide more varied scores. DeepSeek as it is, however, seems to fall behind ChatGPT and Gemini, which provided a varied score distribution similar to human raters, while sustaining high consistency as well. Unlike the general performance of Generative AI staying behind state-of-the-art solutions (Kocoń et al., 2023), these two models have proved themselves as excellent scoring tools in terms of consistency.

One criticism against Generative AI has been their rather controversial interrater reliability with human raters (e.g., Khademi, 2023; Manning et al., 2025). In the current study, Gemini showed an almost excellent agreement with human raters, indicated by an ICC value of 0.735 (Cicchetti, 1994). DeepSeek and ChatGPT exhibited good agreements with 0,695 and 0,673 ICC values, respectively. These measurements were consistent across three measurement pairs, suggesting that Generative AI can indeed provide acceptable interrater reliability. In particular, Gemini performed exceptionally in all three scoring rounds, and considering its high consistency, it showed potential as a reliable essay-scoring tool. Despite its less varied scoring pattern, the similarity of DeepSeek with human raters is also impressive, and with additional improvements, this Generative AI tool can also be a good candidate for carrying out essay-scoring tasks. Although ChatGPT performed worst in terms of both consistency and interrater reliability, its estimates can still be regarded as acceptable in a hybrid scoring system combining human raters and Generative AI. As training reliable human raters for essay scoring necessitates massive amounts of time and resources (Meyer et al., 2023), the promising results of the AI models in this study are crucial, especially for the fields of education, computational linguistics, and corpus linguistics.

The reliability of Generative AI models, however, changed when essays included multiple idioms. Similar to the challenge experienced by second language learners for idiomatic expressions (Beck & Weber, 2016), all three models showed less interrater reliability with human raters when idioms were present, with only Gemini keeping a good reliability level across scoring rounds (Figures 3a and 3b). The findings further support the challenge Generative AI models have with idioms (Olivero, 2024; Sahari et al., 2024), at least for ChatGPT and DeepSeek. Similar to the struggle AI models have in translation tasks with low-resource languages (Hendy et al., 2023; Jiao et al., 2023), their efficiency in processing idiomatic language in essays may also be limited, likely due to the underrepresentation of idioms in their datasets compared to literal language. Alternatively, Generative AI models might fail to capture the right balance in using idiomatic language, either underestimating their value in the right context or overestimating their contribution when used in a repetitive and mechanical way.

The last argument received further support in the current study for the two AI models, ChatGPT and DeepSeek, but not for Gemini. Despite its lower scores overall, Gemini mimicked the human scoring pattern best across different total idiom values; essays with a few idioms received higher scores than idioms with no idioms, but when the number of idioms increased, the scores given by both human raters and Gemini decreased, suggesting a potential penalization for idiom repetition (Figure 4). Delving deeper into the issue, an idiom repetition value was calculated by taking account of repeated idioms in each essay. The results were again similar, and although AI models and human raters penalized idiom repetition, DeepSeek and ChatGPT failed to capture the variety and prominence of the scoring curve created by Gemini and human raters (Figure 5). Gemini also provided the best interrater reliability values when unique idioms were considered, with DeepSeek and ChatGPT showing unacceptable reliability when the number of unique idioms was more than 5.

The overall findings further supported the idea that a hybrid scoring procedure involving both human raters and AI models can bring efficiency and reliability at the same time (e.g., Khademi, 2023; Mizumoto & Eguchi 2023). While Generative AI tools like ChatGPT and DeepSeek can take over the role of a second rater, human raters sustain their essential value by preventing potential AI calculations, especially for essays with multiple idioms, which seem to present an additional challenge due to their figurative nature. Gemini, however, seems to be the best candidate for such a collaboration and for taking over essay-scoring tasks altogether in the future, owing to its ability to process idiom presence and repetition in a similar way to human raters.

**Conclusion**

The three Generative AI models showed promising consistency in scoring essays, but their interrater reliability with human raters was only average-level, except for Gemini. When essays included idioms, Gemini thrived again in mimicking human scoring patterns and sustaining a high-reliability value, outperforming its competitors, ChatGPT and DeepSeek. Reaching human-like efficiency in scoring requires understanding the niceties of idiomatic language as well as the inelegance of idiom repetition in a formulaic way. For that reason, Gemini seems to be the best candidate for integrating AI tools into a hybrid essay-scoring procedure with human raters, and it exhibits remarkable performance in analyzing idiom usage patterns with a level of precision similar to trained human raters.

**TABLES**

Table 1. The results of the GAM model.

| Predictors | AI Score Difference | | |
|---|---|---|---|
| | *Estimates* | *CI* | *p* |
| (Intercept) | 0.32 | 0.21 – 0.43 | **<0.001** |
| word count | -0.00 | -0.00 – -0.00 | **<0.001** |
| ai_deep | -0.77 | -0.85 – -0.68 | **<0.001** |
| ai_gem | -0.72 | -0.81 – -0.64 | **<0.001** |
| ai_gpt | -0.68 | -0.77 – -0.59 | **<0.001** |
| total idioms nor × humanrated | | | **0.015** |
| total idioms nor × ai_deep | | | **0.004** |
| total idioms nor × ai_gem | | | **<0.001** |
| total idioms nor × ai_gpt | | | **0.001** |
| Smooth term (essay id) | | | **<0.001** |
| Observations | 3480 | | |
| $R^2$ | 0.163 | | |

**FIGURES**

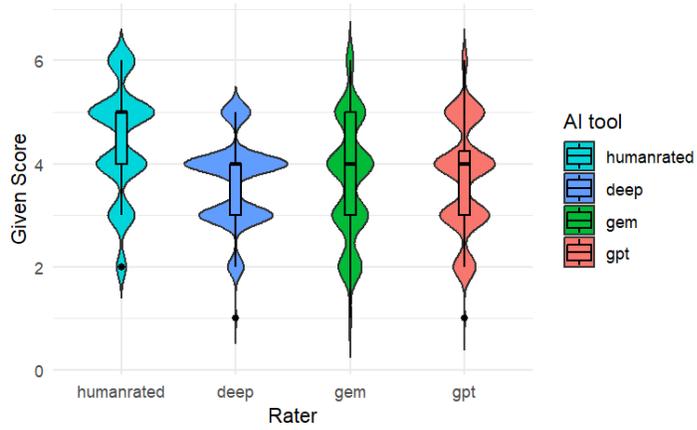

Figure 1. Given scores by human raters and Generative AI.

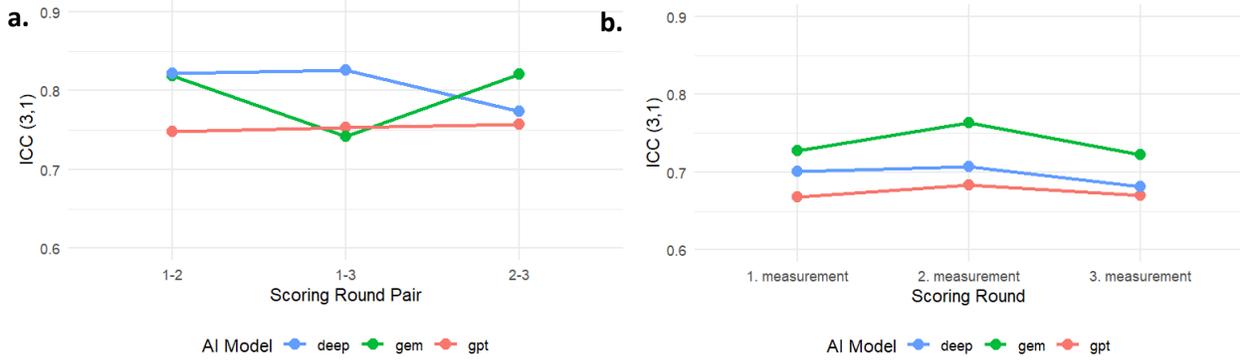

Figure 2. Intra-rater reliability (a) and inter-rater reliability compared to human raters (b) for Generative AI models.

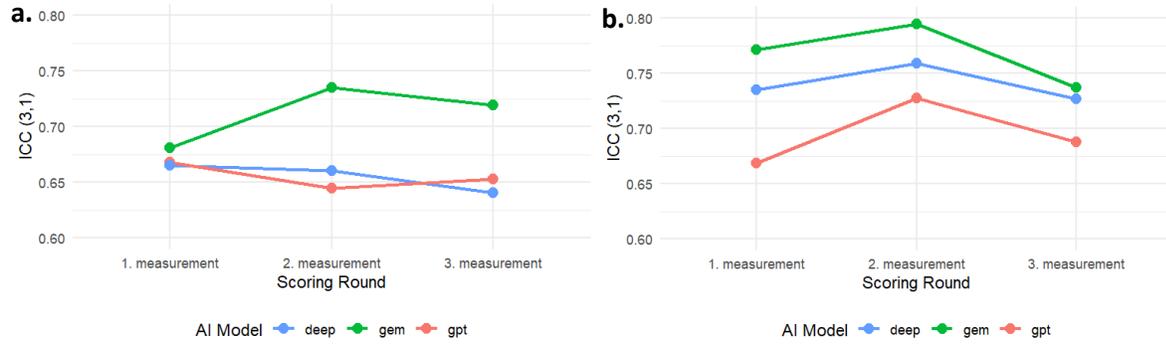

Figure 3. Inter-rater reliability between human raters and AI models for essays with (a) and without (b) idioms.

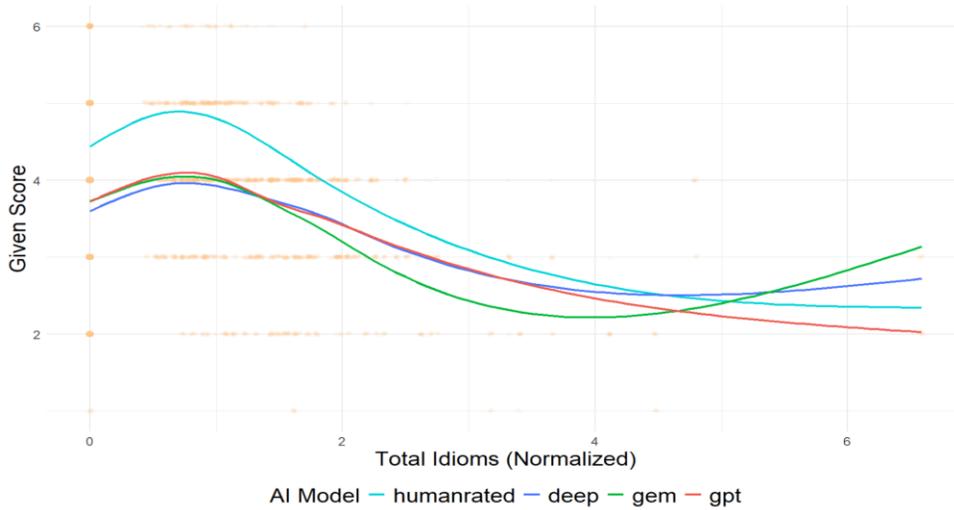

Figure 4. Scoring patterns of human raters and AI models across different idiom numbers.

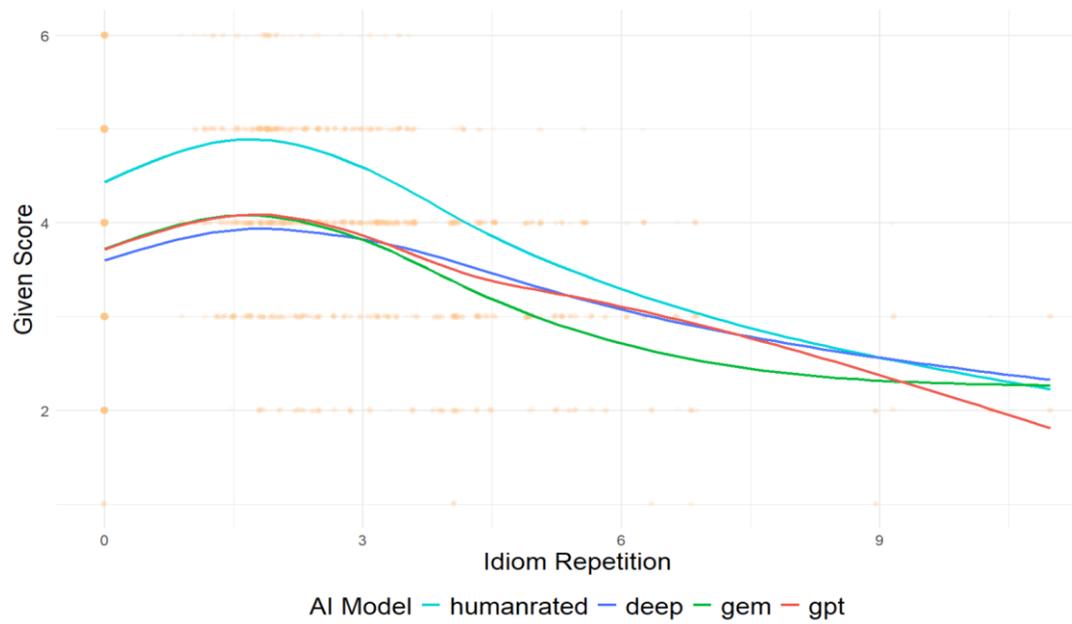

Figure 5. Scoring patterns of human raters and AI models across idiom repetition values.

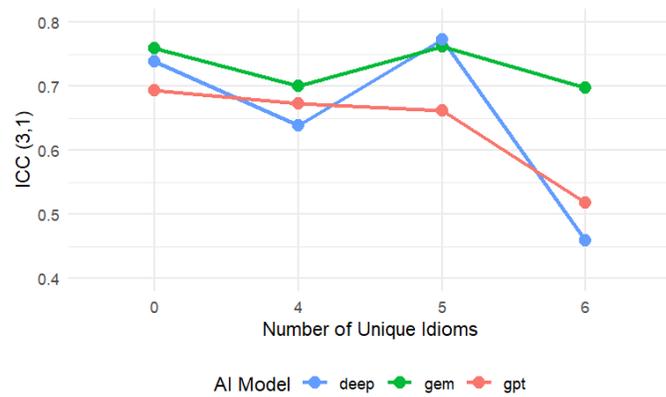

Figure 6. Inter-rater reliability between human raters and AI models across different numbers of unique idioms in essays.